%% file: acl2023.tex
\pdfoutput=1

\documentclass[11pt]{article}

\usepackage{ACL2023}

\usepackage{times}
\usepackage{latexsym}

\usepackage[T1]{fontenc}

\usepackage[utf8]{inputenc}

\usepackage{microtype}

\usepackage{inconsolata}

\usepackage{url}
\usepackage{amsmath}
\usepackage{amsfonts}
\usepackage{amssymb}
\usepackage{graphicx}
\usepackage{color}
\usepackage[none]{hyphenat}
\usepackage{caption}
\usepackage{subcaption}
\usepackage{multirow}
\usepackage{svg}
\usepackage{wrapfig}

\title{No Strong Feelings One Way or Another: Re-operationalizing Neutrality in Natural Language Inference}

\author{Animesh Nighojkar, Antonio Laverghetta Jr., \and John Licato \\
  Advancing Machine and Human Reasoning (AMHR) Lab \\
  University of South Florida\\
  \texttt{\{anighojkar, alaverghett, licato\} @usf.edu} \\}

\begin{document}
\maketitle

\begin{abstract}
Natural Language Inference (NLI) has been a cornerstone task in evaluating language models' inferential reasoning capabilities. However, the standard three-way classification scheme used in NLI has well-known shortcomings in evaluating models' ability to capture the nuances of natural human reasoning. In this paper, we argue that the operationalization of the \textit{neutral} label in current NLI datasets has low validity, is interpreted inconsistently, and that at least one important sense of neutrality is often ignored. We uncover the detrimental impact of these shortcomings, which in some cases leads to annotation datasets that actually \textit{decrease} performance on downstream tasks. We compare approaches of handling annotator disagreement and identify flaws in a recent NLI dataset that designs an annotator study based on a problematic operationalization. Our findings highlight the need for a more refined evaluation framework for NLI, and we hope to spark further discussion and action in the NLP community.
\end{abstract}


\input{latex/introduction.tex}
\input{latex/related.tex}

\input{latex/dissagreement}
\input{latex/neutrals.tex}
\input{latex/unli.tex}

\input{latex/conclusion.tex}

\section*{Limitations}
We approximated the operationalization of the two senses of neutrality using annotator agreement. Perhaps a better basis for operationalizing the two senses of neutrality could be found in the reasons behind the annotators choosing the neutral label. Since no NLI datasets ask annotators to explain their choice and release those responses, we will try to analyze this in the future.

We presented a surface-level syntactic analysis of the differences between the two types of neutrals, but semantic differences should also be analyzed. Intuitively, semantic differences might give us a better understanding of these two types, but further study is needed to verify this.

Though we focus on UNLI as a case study to back up our claims, further analysis on a broader range of NLI datasets (and possible extensions to tasks beyond NLI) should also be conducted.


\bibliography{refs}
\bibliographystyle{acl_natbib}




\end{document}

%% file: latex/introduction.tex
\section{Introduction}\label{sec:introduction}

With the rise of large language models like GPT-3 \cite{brown2020language}, PaLM \cite{palm}, and GPT-4,\footnote{\url{https://openai.com/research/gpt-4}} it has become increasingly necessary to evaluate their language understanding and reasoning abilities. One influential task in this regard is natural language inference (NLI) \cite{maccartney2009extended,MacCartney2014}, which is used to examine the inferential and commonsense reasoning skills of language models \cite{Jeretic_Warstadt_Bhooshan_Williams_2020}. NLI requires a model to determine the relationship between a statement, known as the \textit{premise} $P$, and another statement, called the \textit{hypothesis} $H$, by classifying it as \textit{entailment} (H must be true given P), \textit{contradiction} (H must be false given P), or \textit{neutral} (H can or cannot be true given P).\footnote{Recognizing textual entailment (RTE) \cite{Dagan2006}, a variant of NLI, only considers entailment and non-entailment.} NLI is crucial because it involves comprehending the logical properties of sentences, which is arguably a core capability of human reasoning and an important skill for language models to possess. 

\begin{figure}
    \centering
    \includegraphics[width=0.49\textwidth]{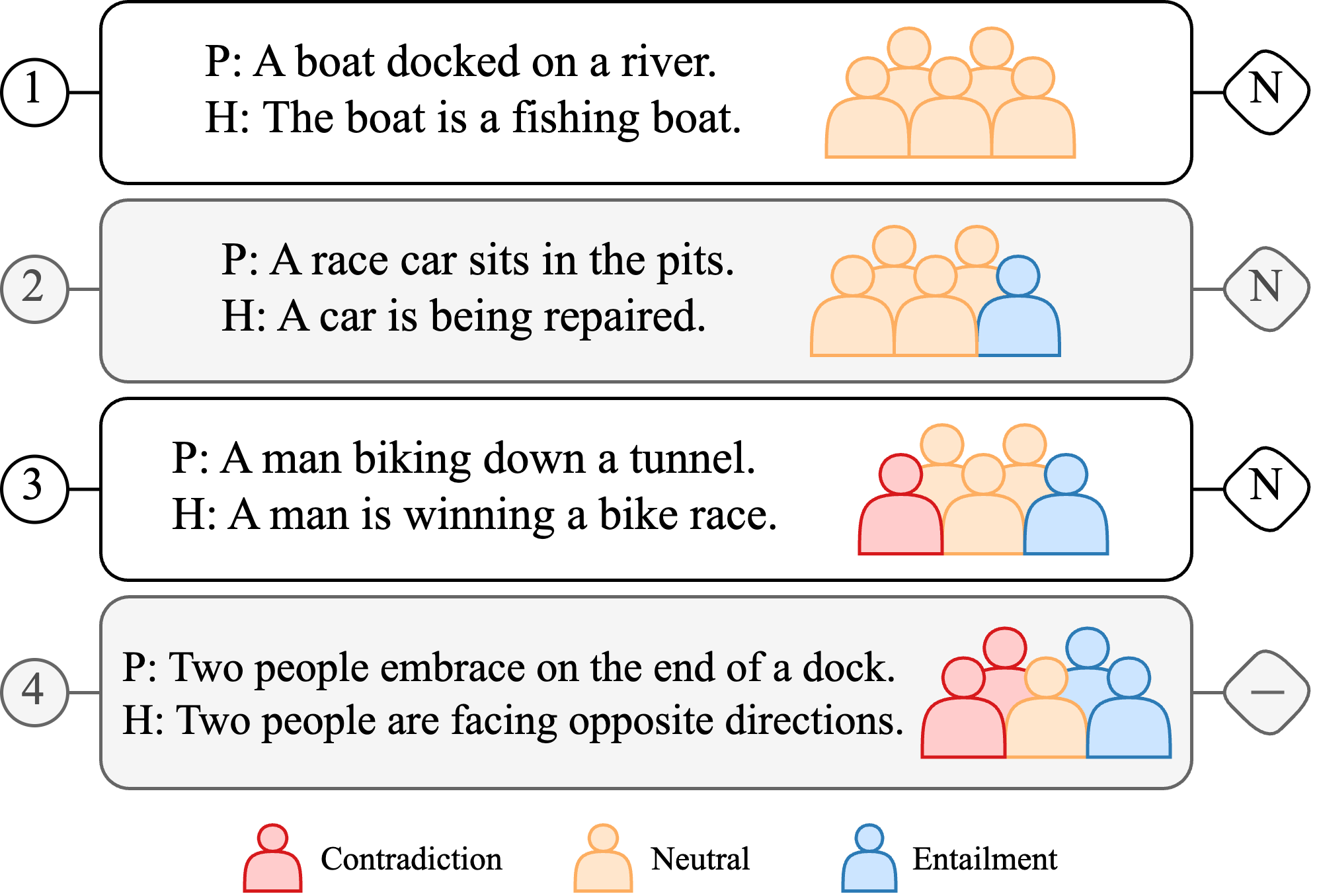}
    \caption{Selected NLI items from SNLI with annotations (shown by colors). The diamonds on the right show the gold label for these items in SNLI; note item 4 is marked `\texttt{-}' and is not assigned a gold label (hence it is ignored). We argue that items with all four annotation distributions should be considered neutral, but that there should be at least two sub-types of neutral.}
    \label{fig:nli}
\end{figure}

Solving NLI requires the ability to perform textual inference between any two sentences (and in some cases, between any two arbitrarily long texts), making it a versatile framework for developing and evaluating reasoning benchmarks. Many NLP tasks, like question answering \cite{Demszky_Guu_Liang_2018}, dialog systems \cite{gong2018natural}, machine translation \cite{Poliak_Belinkov_Glass_Durme_2018}, identifying biased or misleading statements \cite{Nie_Chen_Bansal_2018}, fake news detection \cite{yang2019fake}, paraphrase detection \cite{nighojkar2021improving,nighojkar2021mutual}, and fact verification \cite{Thorne_Vlachos_Christodoulopoulos_Mittal_2018}, require understanding and reasoning about the meaning of text and can be re-framed as NLI problems. NLI provides a broad framework for studying and alleviating logical inconsistencies in a language model's reasoning \cite{poliak-2020-survey,mitchell-etal-2022-enhancing} including explanation-based maieutic prompting \cite{jung2022maieutic}, that uses NLI to evaluate individual links in a reasoning chain.

Most NLI datasets \cite{Bowman_Angeli_Potts_Manning_2015,Williams_Nangia_Bowman_2018,Nie_Williams_Dinan_Bansal_Weston_Kiela_2020,chen-etal-2020-uncertain} utilize crowdsourcing to either generate NLI items or gather labels for pre-existing items. While this approach has advanced research on textual entailment, we believe that current NLI datasets, both established and recent, have overlooked important issues in their annotation design that hinder their validity as measures of textual entailment. Although the effects of different crowdsourcing schemes for NLI dataset development has been studied \cite{bowman-dahl-2021-will, parrish2021does}, we focus on a specific issue: the operationalization of \textit{neutral}. Neutral items usually have the lowest levels of annotator agreement \cite{nie-etal-2020-learn}, and we contend that this disagreement has been handled improperly in previous work, contributing to the ongoing debate about how to handle disagreement in NLI \cite{Palomaki2018,Pavlick_Kwiatkowski_2019,Bowman_Angeli_Potts_Manning_2015,Williams_Nangia_Bowman_2018}. Instructions provided to annotators for labeling items as neutral are often ambiguous and inconsistent between datasets, with phrases like ``neither'' \cite{Nie_Williams_Dinan_Bansal_Weston_Kiela_2020} or ``might be correct'' \cite{Bowman_Angeli_Potts_Manning_2015,Williams_Nangia_Bowman_2018}.

We believe these problems can be addressed by reconsidering the prevailing operationalization of neutral and replacing it with one which embraces disagreement. Although we are not the first to argue for the importance of properly incorporating disagreement \cite{Palomaki2018,Pavlick_Kwiatkowski_2019,basile-etal-2021-need,plank-2022-problem,rottger-etal-2022-two,Uma2022}, we identify specific problems introduced by ignoring disagreement (for example, by dropping examples with low agreement entirely), and offer new evidence supporting its adoption grounded in the psychometric concept of \textit{construct validity}. Consider the items shown in Figure \ref{fig:nli}, sourced from the SNLI dataset \cite{Bowman_Angeli_Potts_Manning_2015}. A general consensus on the gold label is reached by the annotators in the first three items, but the fourth item exhibits a high degree of disagreement. While the first three items are labeled neutral in SNLI and used to train models, the fourth is labeled with a special `\texttt{-}' class, indicating an irresolvable level of disagreement, and hence it is removed from training data \cite{Bowman_Angeli_Potts_Manning_2015}. This practice (also used by \citeauthor{Williams_Nangia_Bowman_2018}) effectively treats disagreement as an undesirable product of NLI data collection---a \textit{linguistic annotation artifact} to be considered as noise rather than signal. But what is the source of this disagreement? Should item 4 in Figure \ref{fig:nli} be ignored, or is it simply a different form of neutrality? We argue that item 4 should be considered a different sense of neutral than the one represented by item 1, because two interpretations are possible: (1) the individuals in the embrace may be facing in opposite directions, resembling a conventional embrace, and (2) one individual may be embracing the other from behind, thereby causing them to face the same direction. This ambiguity in how to interpret such items leads to two irreconcilable types of neutrals; items can be either \textit{true} neutrals (item 1 in Figure \ref{fig:nli}), or they can be neutral as a result of \textit{conflicting} interpretations (item 4). 

\paragraph{Main contributions.} In this paper, we address the aforementioned issues with neutrality in three ways:
\begin{enumerate}
\item We propose a new operationalization of neutral based on inter-annotator agreement, which we argue better captures two distinct senses of neutrality (true neutral and conflicting neutral) often conflated in NLI.
\item We compare our operationalization with a 4-way classification scheme based on annotator disagreement suggested by \citet{Jiang_Marneffe_2019,Zhang_Marneffe_2021,jiang2022investigating} and find that our operationalization has better construct validity, as using it to train models for NLI leads to better downstream performance.
\item We show that known limitations of at least one published NLI dataset (UNLI) are a direct consequence of its adopting an operationalization that did not embrace disagreement, instead opting to aggregate NLI annotations on a continuous scale. We analyze its methodological flaws, and make recommendations to avoid similar problems in future work.
\end{enumerate}


%% file: latex/related.tex
\section{Related Work}\label{sec:related}

NLI is widely used for assessing language models' inferential capabilities, in part due to its generality and versatility. Many datasets, like SNLI \cite{Bowman_Angeli_Potts_Manning_2015}, MultiNLI \cite{Williams_Nangia_Bowman_2018}, Adversarial NLI (ANLI) \cite{Nie_Williams_Dinan_Bansal_Weston_Kiela_2020}, and WA-NLI \cite{liu-etal-2022-wanli} have been developed to evaluate a model's ability to reason through entailment relationships across a wide variety of contexts. Other datasets focus on specific domain knowledge \cite{Holzenberger_Blair-Stanek_Durme_2020,Koreeda_Manning_2021,Yin_Radev_Xiong_2021,Khan_Wang_Poupart_2022,Yang_2022} or require knowledge of non-English languages \cite{Conneau_Rinott_Lample_Williams_Bowman_Schwenk_Stoyanov_2018,Araujo_Carvallo_Kundu_Cañete_Mendoza_Mercer_Bravo-Marquez_Moens_Soto_2022}.

In most NLI datasets, only one label per item is deemed correct, and models are tasked with determining the most plausible of three possible labels. However, there is a growing need for NLI tasks to handle a broader range of relationships and make finer-grained distinctions between them. Researchers are shifting their focus towards finer-grained annotations \cite{chen-etal-2020-uncertain,gantt-etal-2020-natural,Meissner_Thumwanit_Sugawara_Aizawa_2021}, as classical NLI tasks are not well-equipped to handle disagreement between annotators \cite{Zhang_Gong_Choi_2021,Zhang_Marneffe_2021,jiang2022investigating,wang-etal-2022-capture}. Recent research has also focused on assessing models' performance on \textit{ambiguous} NLI items, where humans may disagree on the correct label. ChaosNLI \cite{nie-etal-2020-learn} was developed to study such ambiguities by gathering 100 human annotations on items from a subset of SNLI and MultiNLI, where only 3/5 of annotators agreed on the correct label. They found that models struggled to perform above random chance on items with low inter-annotator agreement and were unable to replicate the annotator label distribution \cite{Zhou_Nie_Bansal_2021}. Since most of the low agreement items are neutral \cite{nie-etal-2020-learn}, we believe a possible reason for this poor performance is the conflation of true and conflicting neutrals as a single category (Section \ref{sec:neutrals}).

\citet{Zhou_Nie_Bansal_2021,Meissner_Thumwanit_Sugawara_Aizawa_2021} build on ChaosNLI and test language models' ability to recover the original annotator label distribution. However, the best results are still below estimated human performance. To solve ambiguous NLI items, \citet{wang-etal-2022-capture} argue that models need to be well-calibrated (i.e., their predicted probability distribution must correctly match the annotator distribution), and they show that label smoothing or temperature scaling can achieve competitive performance without direct training on the label distribution, though it should be noted that other work has found mixed success with using either of these approaches to address ambiguity in NLI \cite{uma2022scaling}. According to \citet{Pavlick_Kwiatkowski_2019}, annotator disagreements are \textit{irresolvable} even when the number of annotators and context are both increased. Such items should not be ignored since the disagreement cannot be always attributed to noise. They argue that handling disagreements should be left to the ones using the models trained on these datasets. Similar to \citet{Zhou_Nie_Bansal_2021}, \citet{Pavlick_Kwiatkowski_2019} also show that NLI models trained to predict one label cannot capture the human annotation distribution.

Despite calls in the literature for annotator disagreement to be accommodated rather than ignored, how this should be done has been the subject of much study. The earliest attempts from SNLI and MultiNLI simply assigned a `\texttt{-}' label to cases that had sufficiently low agreement, indicating that they should not be used for training \cite{Bowman_Angeli_Potts_Manning_2015,Williams_Nangia_Bowman_2018}. More recent work has tried to incorporate low agreement items as a fourth \textit{disagreement} class, a practice that began with \citet{Jiang_Marneffe_2019} and was later used by \citet{Zhang_Marneffe_2021, jiang2022investigating}. We examine this practice in Section \ref{sec:dissagreement} and demonstrate that simply using a \textit{catch-all} category for disagreement is not as effective as our operationalization for neutral items.

Another line of research has explored changing the annotation schema to use a continuous scale, rather than a discrete one, in the hope that this type of scale will better capture the subtleties of reasoning over ambiguity and lead to less disagreement. \citet{chen-etal-2020-uncertain} introduce \textit{uncertain natural language inference} (UNLI), where annotators indicate the likelihood of a hypothesis being true given a premise. While models trained on UNLI can closely predict human estimations, later work has found that fine-tuning on UNLI can hurt downstream performance \cite{Meissner_Thumwanit_Sugawara_Aizawa_2021}, suggesting a serious flaw in the UNLI dataset. 
We analyze further issues with UNLI in Section \ref{sec:unli}.

In a recent study, \citet{Kalouli_Hu_Webb_Moss_Paiva_2023} propose a new interpretation of neutral based on the concept of \textit{strictness}. They argue that, under ``strict interpretation'', the pair \textit{P: The woman is cutting a tomato. H: The woman is slicing a tomato/} would be considered neutral as she could be cutting squares, but it could be considered an entailment pair if the interpretation is not so strict. Their operationalization of neutral based on the concept of \textit{strictness} lacks clarity due to the absence of a precise, understandable definition of \textit{strictness}. In effect, it simply shifts the problem of understanding what makes a pair of sentences neutral to understanding what makes their relationship ``strictly logical'' (a term they use to define strict interpretation, without further elaboration).\footnote{Note that the strict conditional $\square (p \rightarrow h)$ was famously introduced by \citet{Lewis1912} as a formalization of the indicative conditional. However, this does not appear to be the sense of ``strict'' meant by \citet{Kalouli_Hu_Webb_Moss_Paiva_2023}.} 

%% file: latex/dissagreement.tex
\section{Empirical evaluation of `disagreement' as a fourth class} \label{sec:dissagreement}

\begin{figure*}
    \centering
    \begin{subfigure}[b]{0.53\textwidth}
        \includegraphics[width=\textwidth]{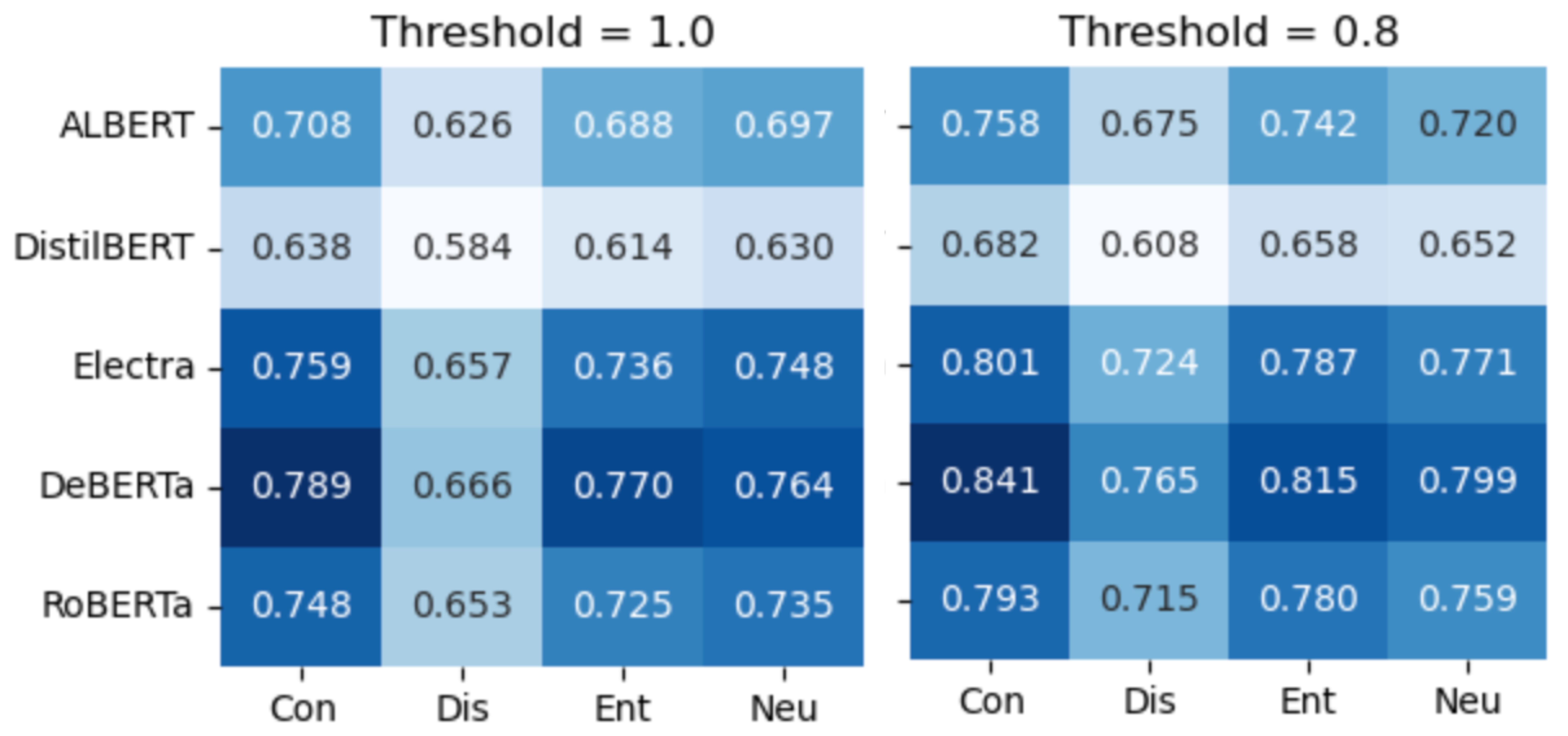}
        \caption{MultiNLI}
        \label{fig:heatmap-mnli}
    \end{subfigure}
    \hfill
    \begin{subfigure}[b]{0.46\textwidth}
        \includegraphics[width=\textwidth]{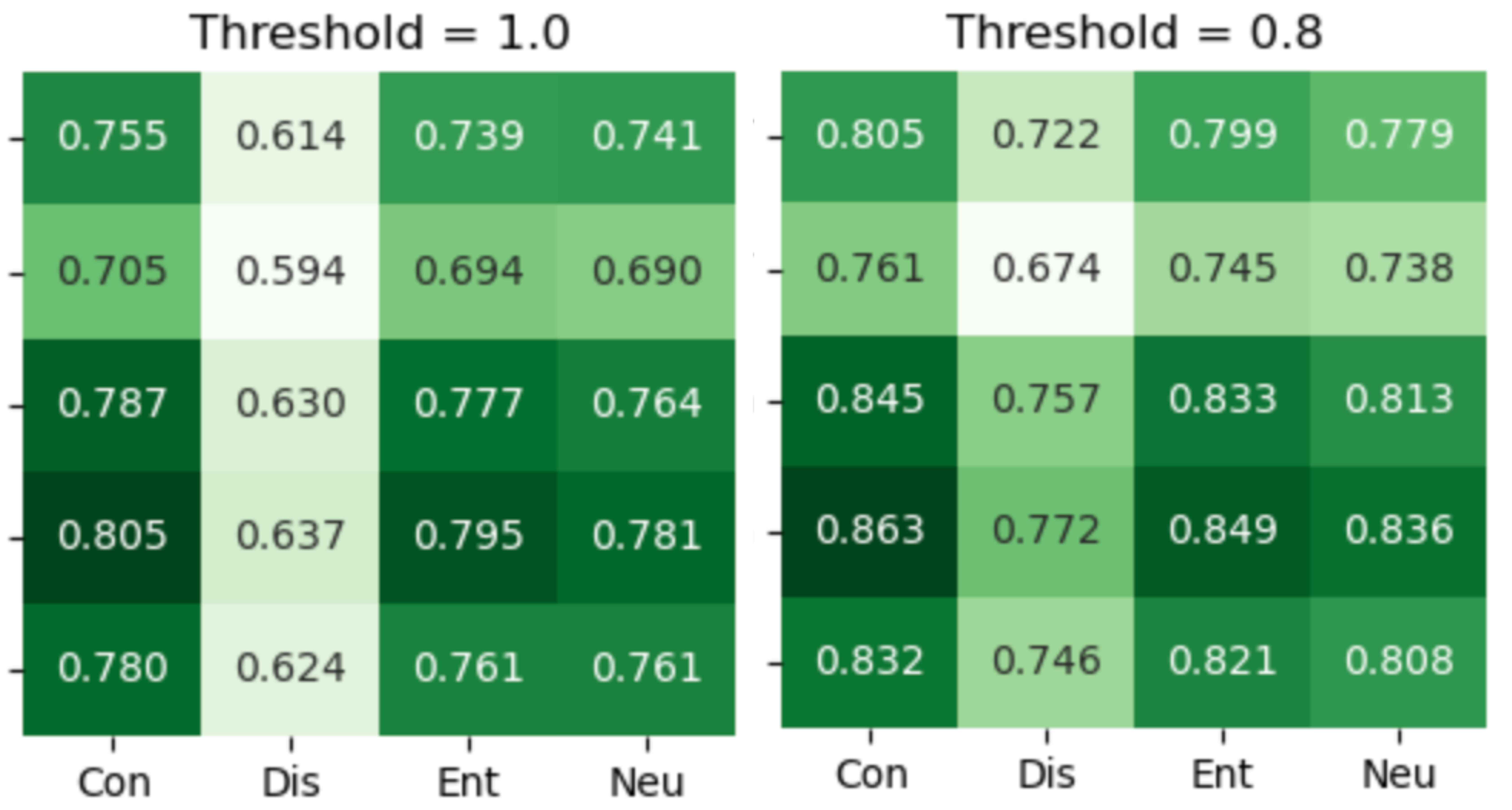}
        \caption{SNLI}
        \label{fig:heatmap-snli}
    \end{subfigure}
    \caption{Heatmaps of $F_1$ scores on different 4-way classification schemes (x-axis) for different language models (y-axis). Darker boxes indicate better performance. Models consistently under-perform on the disagreement-based classification scheme (\textbf{Dis}) proposed by \citet{Jiang_Marneffe_2019,Zhang_Marneffe_2021,jiang2022investigating}, indicating that a catch-all disagreement label does not provide enough information to models to reason over ambiguous items.}
    \label{fig:heatmaps}
\end{figure*}


The classification scheme that uses a fourth `disagreement' label for low-agreement items \cite{Jiang_Marneffe_2019,Zhang_Marneffe_2021,jiang2022investigating} conflates all three NLI labels in doing so. To explore this possibility, we conduct an empirical study to compare this disagreement-based scheme with other 4-way classification schemes. We define the \textit{level of agreement} ($\mathbf{A}$) between annotators on NLI items as:
\begin{equation}
    \mathbf{A} = \frac{\textit{number of votes for the majority label}}{\textit{total number of votes}}
    \label{eq:A}
\end{equation} 
We also explore two agreement threshold $t$ values ($0.8$, and $1$),\footnote{Because SNLI and MultiNLI have at most 5 annotations, and the majority label is always taken as the gold label, $0.4$ is the smallest possible $\mathbf{A}$ that can be used. Since all items at that agreement are marked as \texttt{-} in both the datasets, $t=0.6$ cannot be used for \textbf{Ent} and \textbf{Con}. Also, $t=0.6$ will give us same items for all four classes in \textbf{Dis} as well as \textbf{Neu}, making their comparison at that threshold meaningless.} which is the cutoff-value of $\mathbf{A}$ below which items are considered to have ``low agreement.'' Note that \citet{Jiang_Marneffe_2019} choose $t=0.8$ but do not provide an explanation for choosing it. We train ALBERT-base \cite{lan2019albert}, DistilBERT-base-uncased \cite{sanh2019distilbert}, Electra-base \cite{electra}, DeBERTa-v3-base \cite{he2020deberta}, and RoBERTa-base \cite{Liu_Ott_Goyal_Du_Joshi_Chen_Levy_Lewis_Zettlemoyer_Stoyanov_2019} to show that these results are not specific to just a few models. We are limited to using SNLI and MultiNLI because they are the only NLI datasets that report individual annotations in sufficient quantity to finetune transformer language models. We trained each model for 5 epochs and tested their performance on a held out, stratified, evaluation set.\footnote{Github code will be released upon publication.}. We use only the base versions of these models because our objective here is not to train the best models, but to examine and compare classification schemes. Models are being used in this experiment only to compare the \textit{separability} of all classes for each of these classification schemes:
\begin{itemize}
    \item \textbf{Con:} Entailment, Neutral, $\uparrow$ Contradiction, $\downarrow$ Contradiction \footnote{$\uparrow$ and $\downarrow$ denote high and low annotator agreement respectively.}
    \item \textbf{Dis:} Entailment, Neutral, Contradiction, Disagreement
    \item \textbf{Ent:} $\uparrow$ Entailment, $\downarrow$ Entailment, Neutral, Contradiction
    \item \textbf{Neu:} Entailment, $\uparrow$ Neutral, $\downarrow$ Neutral, Contradiction
\end{itemize}
Better F$_1$ scores would suggest the model could better differentiate between the classes of the given classification scheme, and thus the scheme has better \textit{ecological validity}.\footnote{Ecological validity examines whether the results of a study can be generalized to real-life settings \cite{egger2008systematic}.}

Results are shown in Figure \ref{fig:heatmaps}. We find that using a fourth `disagreement' label leads to the worst results consistently. These results suggest that having a catch-all `disagreement' label does not provide enough information to help models successfully reason over ambiguous items. Note that unlike the other three schemes, \textbf{Dis} classifies all low-agreement items as `disagreement', thus making the other three schemes more imbalanced than \textbf{Dis}. For instance, \textbf{Con} classifies only low-agreement contradiction items as the fourth class and low-agreement neutral and entailment items are classified as their respective majority labels. Lowest $F_1$ score on \textbf{Dis} (the most balanced classification scheme) is perhaps even more informative than it would have been if the schemes were equally balanced. Any of the other three schemes consistently leads to better performance, regardless of model or threshold used, and thus has better construct validity \cite{bleidorn2019using,zhai2021validity} than the classification scheme based on disagreement.






%% file: latex/neutrals.tex
\section{Operationalizing Neutral}\label{sec:neutrals}


In NLI, the neutral label is used for situations where the relationship between the premise and hypothesis is ambiguous or there is insufficient information to determine the relationship. Neutral is often considered a catch-all for relationships that do not fall under entailment or contradiction. The definition of neutral is typically provided to crowd-source workers as ``neither'' \cite{Nie_Williams_Dinan_Bansal_Weston_Kiela_2020} or ``might be correct'' \cite{Bowman_Angeli_Potts_Manning_2015,Williams_Nangia_Bowman_2018}. 

But is a classification of neutral simply a default assumption that always means neither entailment nor contradiction can be definitively determined, or can it be a positive claim that a different type of relationship holds between the sentences? A closer look at the data obtained from NLI datasets suggests that neutrality is more complex than it may initially seem. According to \citet{nie-etal-2020-learn}, neutral items in many NLI datasets exhibit the lowest agreement levels. The most frequent label below an agreement level of $\mathbf{A}=0.8$ for both the SNLI and MultiNLI subsets is neutral, while it is the least frequent label at a perfect agreement level. This lack of agreement motivates our focus on neutral particularly, as it is consistently the most problematic label to annotate. The empirical study in Section \ref{sec:dissagreement} also shows that a neutral-based classification scheme has a better separability than a disagreement-based classification scheme.

There are at least two senses in which the relationship between two sentences can be said to be neutral, which become clear if we imagine two possible justifications that an individual NLI annotator may provide for why they selected the label neutral: (1) \textit{True Neutral:} The annotator cannot find any sufficiently strong reasons (using whichever standard of strength they determine appropriate) to satisfy either entailment or contradiction; or (2) \textit{Conflicting Neutral:} The annotator finds strong reasons to support \textit{both} entailment and contradiction. 

It is a central position of this paper that these two interpretations of the neutral label are irreconcilable and should not be confused with each other. Attempting to conflate the two, e.g.\ by assuming that neutrality is simply the mid-point on a continuous scale between the two extremes of entailment and contradiction, will and has led to significant reductions in quality of data collections and their resulting benchmark datasets (see \S \ref{sec:unli}).

\begin{table*}
    \footnotesize
    \centering
    \begin{tabular}{|c|c|c|c|c|}
        \hline
        Dataset & Mean Length ($T$) & Mean Length ($C$) & Reading Ease ($T$) & Reading Ease ($C$) \\
        \hline
        $*$ SNLI dev + test & 109.6 & 118.2 & 84.0 & 82.8 \\
        SNLI train & 102.8 & 111.3 & 84.8 & 83.6 \\
        $*$ MultiNLI matched + mismatched & 172.0 & 183.0 & 67.0 & 65.2 \\
        MultiNLI train & 163.8 & 186.0 & 68.7 & 64.4 \\
        ANLI R3 dev & \textcolor{brown}{\underline{389.0}} & \textcolor{brown}{\underline{372.7}} & \textcolor{brown}{\underline{67.9}} & \textcolor{brown}{\underline{65.3}} \\
        ANLI R3 test & 382.4 & 392.7 & \textcolor{brown}{\underline{69.8}} & \textcolor{brown}{\underline{66.1}} \\
        ANLI R3 train & 369.3 & 377.3 & \textcolor{brown}{\underline{66.3}} & \textcolor{brown}{\underline{64.6}} \\
        WA-NLI test & 147.3 & 147.6 & \textcolor{brown}{\underline{77.4}} & \textcolor{brown}{\underline{77.4}} \\
        WA-NLI train & 147.5 & 148.6 & 77.1 & 77.0 \\
        \hline
    \end{tabular}
    \caption{Comparison of true ($T$) and conflicting ($C$) neutrals. Smaller values for reading ease indicate harder-to-read items. We use our trained model to estimate $\mathbf{A}$ for the datasets that do not release individual annotations and the ones that do are marked with a ``$*$''. Cases where our hypothesis was NOT confirmed are underlined and in brown.}
    \label{tab:linguistic_analysis}
\end{table*}

No existing NLI dataset, to our knowledge, asks or encourages annotators to explain whether their reasons for selecting neutral are in line with true or conflicting neutral as we have defined them above. For the present work, then, we present evidence for the discriminant validity of true and conflicting neutral (i.e., that they refer to two distinct constructs that can and should be measured separately \citet{campbell1959convergent}) by assuming that they will be \textit{approximately reflected in the distribution} of individual annotations on a single NLI item---in other words, conflicting neutral items will tend to have annotation distributions resembling item 4 in Figure \ref{fig:nli}, whereas true neutrals will tend to match item 1. Results in Section \ref{sec:dissagreement} show that indeed such a classification scheme does a much better job of separating the four classes for models than a scheme that conflates all three labels. 

\input{latex/method.tex}

%% file: latex/method.tex
\paragraph{True vs.\ Conflicting Neutral:
Surface-level Differences}

We perform an exploratory analysis to identify potential reasons why annotators may disagree on some `neutral' items, to better motivate our operationalization of `neutral'. Drawing from \citet{Pavlick_Kwiatkowski_2019}, who found that disagreement increases as more context is given, we investigate whether ambiguity in NLI items arises due to increased complexity, leading to difficulties in accurately interpreting them. We measure this complexity using two metrics: mean length of the item in terms of number of characters (after the premise and hypothesis are joined with a space), and Flesch Reading Ease \cite{Flesch_1948}, a commonly-used measure of text readability.
Our findings, shown in Table \ref{tab:linguistic_analysis}, reveal that true neutral items are shorter and easier to read than conflicting neutral items. However, the observed difference in complexity between the two forms of neutrals is marginal and inconclusive. These results suggest that at least superficial qualitative differences exist between different types of neutrals, but more extensive research is needed to clarify the extent of these differences.

%% file: latex/unli.tex
\section{An Analysis of UNLI}\label{sec:unli}

We have argued that a carefully grounded operationalization of the neutral label is crucial for ensuring the reliability (performance should be free from random error) and validity of NLI. To demonstrate the issues that can arise if this caution is not taken, we next analyze a recent NLI dataset --- Uncertain NLI (UNLI) \cite{chen-etal-2020-uncertain}. The UNLI dataset, when used for fine-tuning, appears to actually harm downstream performance \cite{Meissner_Thumwanit_Sugawara_Aizawa_2021,Zhou_Nie_Bansal_2021,wang-etal-2022-capture}. UNLI attempts to enhance NLI by converting the categorical labels for some SNLI items to a continuous scale. Participants were instructed to rate the likelihood of a given hypothesis being entailed by a given premise using an ungraduated slider, ranging from 0 (labeled as ``impossible'') to 1 (labeled as ``very likely'') and were shown the probability they were assigning to the \textit{premise-hypothesis} pair in real time.

According to \citet{chen-etal-2020-uncertain}, the probabilistic nature of NLI \cite{glickman2005} suggests that not all contradictions or entailments are equally strong.\footnote{The view that NLI is inherently probabilistic, or that natural inference can be best modeled with probability, is not universally held, e.g. \cite{Bringsjord2008c}.} Thus, UNLI was developed with the intention of capturing subtler distinctions in \textit{entailment strength} using a continuous scale. This dataset has over 60K items from SNLI, annotated by humans. For each premise-hypothesis pair, two annotations were collected, and in cases where the first two annotators differed by $20\%$ or more, a third annotator was consulted. However, the dataset only reports the averaged scores, which makes it impossible to assess the degree of agreement or correlation between the two annotators or even identify examples where a third annotator was needed. Thus, reported values near 0.5 (which we might take to be the equivalent of \textit{neutral} items) fundamentally conflate items where both annotators chose the midpoint on the slider with items where each annotator chose one of the extremities. 

The assumption that one continuous scale can capture even the three categories in standard NLI (entailment, contradiction, and neutral) is a strong one (already shown to be problematic in \cite{Pavlick_Kwiatkowski_2019}), which is typically glossed over by presuming that entailment lies at the higher end of the spectrum, contradiction at the other end, and neutral somewhere in the middle. 
But no such instruction to interpret the scale this way was provided to annotators. 
Indeed, as we will show, annotators appeared to be confused as to whether an absence of entailment meant that the slider should be at the `$0$' position, or in the middle.

\begin{figure}
    \centering
    \includegraphics[width=0.49\textwidth]{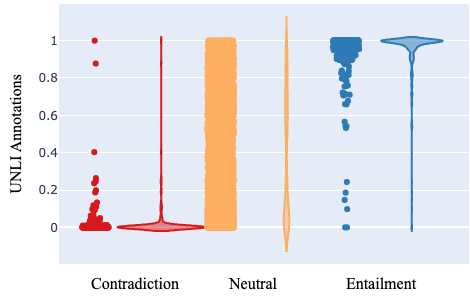}
    \caption{Figure 1 from \citet{chen-etal-2020-uncertain} redrawn on a linear scale. Note the two distinct bulges in the violin plot for neutral items, suggesting that annotators were confused about whether neutral items should be placed near $0$ or middle of the slider.}
    \label{fig:unli}
\end{figure}

\begin{table*}
    \footnotesize
    \centering
    \renewcommand{\arraystretch}{1.1} 
    \begin{tabular}{| p{.45\linewidth} | p{.28\linewidth} | p{.12\linewidth} | p{.04\linewidth} |}
            \hline
            Premise & Hypothesis & SNLI Annotations & UNLI score\\
            \hline
            A woman with a blue jacket around her waist is sitting on the ledge of some stone ruins resting. & A man sits on a ledge. & $4C-0N-1E$ & $0.88$ \\
            A lady is standing up holding a lamp that is turned on. & She is lighting a dark room. & $2C-2N-1E$ & $0.78$ \\
            A singer wearing a leather jacket performs on stage with dramatic lighting behind him. & A singer is on American idol. & $1C-4N-0E$ & $0.01$ \\
            A small boy wearing a blue shirt plays in the kiddie pool. & Boy cooling off during the summer. & $1C-4N-0E$ & $0.89$ \\
            \hline
        \end{tabular}
        \caption{Items from UNLI along with their individual annotations from SNLI.}
        \label{tab:unli}
\end{table*}

In their attempt to obtain subjective probabilities for premise-hypothesis pairs, the authors used a scale with 10K steps with a scaled logistic transformation ($f(x)=\sigma(\beta(x-5000))$) to convert the values on the scale into probabilities between $0$ and $1$. They do not report the chosen value of $\beta$ and do not specify whether the scores were averaged before or after applying the function, which is crucial information as both would yield different results. Because raw values of $x$ are not provided, and we do not know whether scaling is performed before or after averaging, we are unable to recover the chosen values of $\beta$.

The scale \citet{chen-etal-2020-uncertain} used was based on EASL \cite{sakaguchi2018efficient}, an approach developed to collect scalar ratings in NLP tasks.\footnote{This scale was not validated for NLI by \citet{sakaguchi2018efficient} and the tasks they evaluated it for --- like evaluating quality of machine translations, or the frequency of words in language --- differ significantly from NLI.} They then modified the EASL scale by utilizing the aforementioned logistic transformation, which they argued would allow for more nuanced values near both extremes. Notably, the source of the anchor points used on the scale (i.e., ``impossible'' and ``very likely'') is not explicitly stated by \citet{chen-etal-2020-uncertain}, although it is possible they were obtained from JOCI \cite{zhang-etal-2017-ordinal}, a dataset created for studying ordinal commonsense reasoning that uses the same anchor points for opposite ends of the scale.\footnote{This is further supported by the fact that \citet{chen-etal-2020-uncertain} cite \citet{zhang-etal-2017-ordinal} as a previous attempt to model likelihood judgments in NLI, which is also the aim of UNLI.}


In effect, their logistic transformation compresses the extreme ends of the scale, so that the graphic they display (Figure 1 in \citet{chen-etal-2020-uncertain}), at first glance, appears as if the NLI items labeled as contradiction, neutral, and entailment occupy roughly equal space across the continuum of values. Figure \ref{fig:unli} instead depicts the distribution of averaged human responses collected by \citet{chen-etal-2020-uncertain} on a linear scale.\footnote{Many of the properties of the scale we address here were unclear from reading the original figure in \citet{chen-etal-2020-uncertain}, necessitating the redrawing.} It is clear to observe in Figure \ref{fig:unli} that while entailment and contradiction annotations are distinctly separated and skewed heavily towards the extreme opposite ends of the scale, annotations for neutral \textit{span the entire range from 0 to 1}. The origin of this discrepancy is unclear, but based on the instructions given to them, it may be that annotators were unsure where to place neutral on the scale. Supporting this hypothesis is the bulge near $0$ on the violin plot for neutral in Figure \ref{fig:unli}, which suggests that annotators chose $0$ for both neutral and contradiction items. This information is obscured by the logistically transformed graph displayed by \citet{chen-etal-2020-uncertain}.

Table \ref{tab:unli} highlights some examples from UNLI that demonstrate the poor alignment of its annotations with SNLI annotation distributions. From Figure \ref{fig:unli}, the reliability of the scale for neutral annotations is notably poor, with annotations spanning the entire range of the scale. This suggests that neutral annotations lack internal consistency, an important measure of reliability \cite{rust2014modern}, because annotators do not label the NLI items in a consistent fashion even when the label remains constant.

Measurement issues are not uncommon in other fields that routinely run human studies, including psychological and educational mesurement. Development of annotation schemes in these fields often involves careful consideration of the item format, including the rating scale, to ensure that it effectively measures the construct of interest \cite{alma99379871487606599}. This can be achieved through qualitative analysis, such as cognitive interviews and focus groups, where items are administered to test takers and feedback is collected to ensure that the scale is understood and completed accurately, among other things \cite{alma99380256053506599}. However, in the development of UNLI, \citet{chen-etal-2020-uncertain} did not report using such procedures. Moreover, common practices in measurement research were missing from UNLI, such as reporting how bad-faith responses were identified and filtered out, using attention-check items (except the qualifying test, whose results are not provided as part of the dataset), employing a sufficienlty large sample size of annotators, and providing individual annotations and relevant information about the annotators like their recruitment and compensation. These omissions make precise scientific replication impossible, and raise concerns about the validity of UNLI as a measure of (and benchmark for) NLI, while also providing a plausible explanation for why prior research yielded poor results when using UNLI for fine-tuning.


%% file: latex/conclusion.tex
\section{Conclusion}

In this paper, we examined the operationalization of neutral in NLI datasets. Our analysis revealed that previous attempts to handle ambiguity in NLI based on neutrality have significant issues with their validity as annotation strategies for NLI. We proposed a new operationalization of neutral into \textit{true neutral} and \textit{conflicting neutral}. Although instances of these forms of neutral are present in most popular NLI datasets, they have been conflated into one neutral label, limiting our ability to measure ambiguity in NLI effectively. We showed that this approach of casting NLI to a 4-way classification task is better than the disagreement-based classification scheme used in previous work. We used UNLI as a case study to highlight measurement and annotation issues that should be avoided in the future.

Of the many factors that make science successful, two of the most important are the ability to make carefully designed measurements, and replicability. The first of these cannot be met when measurements of constructs are made in ways that significantly compromise their validity and reliability. And replicability is made impossible when papers are published in reputable venues reporting unclear collection details, having important parameter choices omitted, and with datasets reporting summary statistics in place of crucially important data. A significant roadblock of the work we reported in this paper was the lack of availability of individual annotations in widely-adopted NLI benchmarks, even when there seems to be no public benefit in leaving out such information. It is our hope that the present work will encourage our fellow AI researchers to more highly value such considerations.